\documentclass{article}
\usepackage{spconf,amsmath,graphicx,hyperref}
\usepackage{algorithm}
\usepackage{algpseudocode}
\usepackage{varwidth} % 用于处理长行的Require/Ensure

\usepackage{booktabs}
\usepackage{multirow}
\usepackage{graphicx}
\usepackage[table]{xcolor}
\usepackage{colortbl}
\usepackage{enumitem}
\usepackage{amsfonts}
\title{SemantiCache: Efficient KV Cache Compression via Semantic Chunking and Clustered Merging}

\name{\begin{tabular}[t]{c} % [t] 表示顶部对齐
      Shunlong Wu$^{1}$, Hai Lin$^{1,2}$, Shaoshen Chen$^{1}$, Tingwei Lu$^{1}$, Yongqin Zeng$^{1}$, Shaoxiong Zhan$^{1}$,\\
      {Hai-Tao Zheng}$^{1,2\dagger}$\thanks{\textsuperscript{\textdagger} Corresponding author. E-mail: zheng.haitao@sz.tsinghua.edu.cn. This research is supported by National Natural Science Foundation of China (Grant No.62276154);
the Natural Science Foundation of Guangdong Province (Grant No.2024TQ08X729);
Basic Research Fund of Shenzhen City (Grant No.JCYJ20240813112009013 and GJHZ20240218113603006);
The Major Key Project of PCL for Experiments and Applications (Grant No.PCL2024A08).}, {Hong-Gee Kim}$^{3}$
      \end{tabular}}

\address{$^{1}$ Shenzhen International Graduate School, Tsinghua University, Shenzhen, China \\
$^{2}$ Peng Cheng Laboratory, Shenzhen, China 
 \qquad $^{3}$ Seoul National University, South Korea\\}
 
\begin{document}
% \ninept
%
\maketitle
\begin{abstract}

Existing KV cache compression methods generally operate on discrete tokens or non-semantic chunks. However, such approaches often lead to semantic fragmentation, where linguistically coherent units are disrupted, causing irreversible information loss and degradation in model performance. To address this, we introduce \textbf{SemantiCache}, a novel compression framework that preserves semantic integrity by aligning the compression process with the semantic hierarchical nature of language. Specifically, we first partition the cache into semantically coherent chunks by delimiters, which are natural semantic boundaries. Within each chunk, we introduce a computationally efficient \textbf{Greedy Seed-Based Clustering (GSC)} algorithm to group tokens into semantic clusters. These clusters are further merged into semantic cores, enhanced by a \textbf{Proportional Attention} mechanism that rebalances the reduced attention contributions of the merged tokens. Extensive experiments across diverse benchmarks and models demonstrate that SemantiCache accelerates the decoding stage of inference by up to 2.61$\times$ and substantially reduces memory footprint, while maintaining performance comparable to the original model.

\end{abstract}
\begin{keywords}
Large Language Models, KV Cache Compression, Efficient Inference
\end{keywords}

\section{Introduction}
\label{sec:intro}

While the ability of Large Language Models (LLMs) to process diverse tasks is rapidly advancing~\cite{zhao2025cosoptimaleventscheduling,lv-etal-2025-raise,lu2026tokenlineenhancingcode}, this progress is fundamentally constrained by the high cost of autoregressive inference in long context scenarios~\cite{tang2024perception,chen2025dastcontextawarecompressionllms,tang2025gmsa,tang2026comicoarsetofinecontextcompression,tang2026readhumancompressingcontext,lv2026datadistributionmattersdatacentric,zhao2026cometcollaborativememorytransformer,li2025admtree}. A critical bottleneck is the Key-Value (KV) cache, which stores attention keys and values for all preceding tokens. This cache grows linearly with the input sequence, leading to two major challenges: (1) a prohibitive memory footprint that can exhaust GPU resources, and (2) a substantial increase in inference latency.

To address the aforementioned challenges, a range of KV cache compression techniques have been proposed. These methods can be broadly categorized into eviction-based \cite{sepLLM2025,li2024snapkv,xiao2023efficient,zhang2023h2o,ding2025dynamicattention,lv2025kvpruner} and merging-based \cite{wan2024d2o,zhang2024cam,yuan2025weightedkv,wang2024kvmerger}. Eviction-based approaches seek to permanently discard tokens deemed less "critical" based on factors such as attention scores. However, this strategy often results in notable performance degradation due to the irrecoverable loss of information contained within the evicted tokens. To address this limitation, merging-based approaches aim to compress the cache by merging tokens instead of pruning them. 

% While these methods can better preserve information, they typically depend on preliminary token eviction or inefficient nearest-neighbor matching, which constrain their ability to fully retain model performance.

Despite their differences, these existing methods share a critical flaw: they operate on discrete tokens or arbitrary, non-semantic chunks, ignoring the natural structure of language. This approach leads to semantic fragmentation, which breaks apart coherent semantic structures such as phrases, clauses, or sentences. Such fragmentation causes an irreversible loss of critical semantic information, degrading model performance.

To address this issue, we draw inspiration from a linguistic principle: natural language is not a flat stream of words, but a hierarchy of semantic structures \cite{Ding2016}. Our method, SemantiCache, aligns its compression process with this hierarchy to preserve coherent semantic structures and prevent fragmentation. The process operationalizes the cognitive strategy humans employ to efficiently memorize long texts. Specifically, we emulate the hierarchical decomposition of information, where a reader first segments the text into complete semantic units (e.g., sentences and clauses), and then distills the core meaning of each unit by identifying and consolidating its central concepts. SemantiCache mirrors this by first partitioning the cache along natural semantic boundaries to create semantically coherent chunks. Within these chunks, we then analyze their internal structure by introducing a computationally efficient GSC algorithm to group semantically related tokens. These resulting clusters are then merged into compact semantic cores. To counteract the dilution of influence caused by this merging, we further apply a Proportional Attention mechanism, which reweights the attention contribution of each core. This entire process effectively compresses the KV cache while maintaining semantic fidelity. We conduct experiments on various long-context tasks, showing that SemantiCache can accelerate inference decoding by up to 2.61$\times$ and substantially lower the memory footprint, outperforming both eviction-based and merging-based baselines while maintaining performance comparable to the full model.

Our contributions are summarized as follows: (1) We identify and address the prevalent issue of "semantic fragmentation" in existing KV cache compression methods. (2) We introduce a novel compression framework that emulates the human cognitive strategy for memorizing long texts. (3) We design a lightweight clustering algorithm (GSC) for efficient semantic aggregation and apply a Proportional Attention mechanism to address the information dilution problem common in traditional merging methods. 

\begin{figure}[htb]
  \centering
  \includegraphics[width=8.5cm]{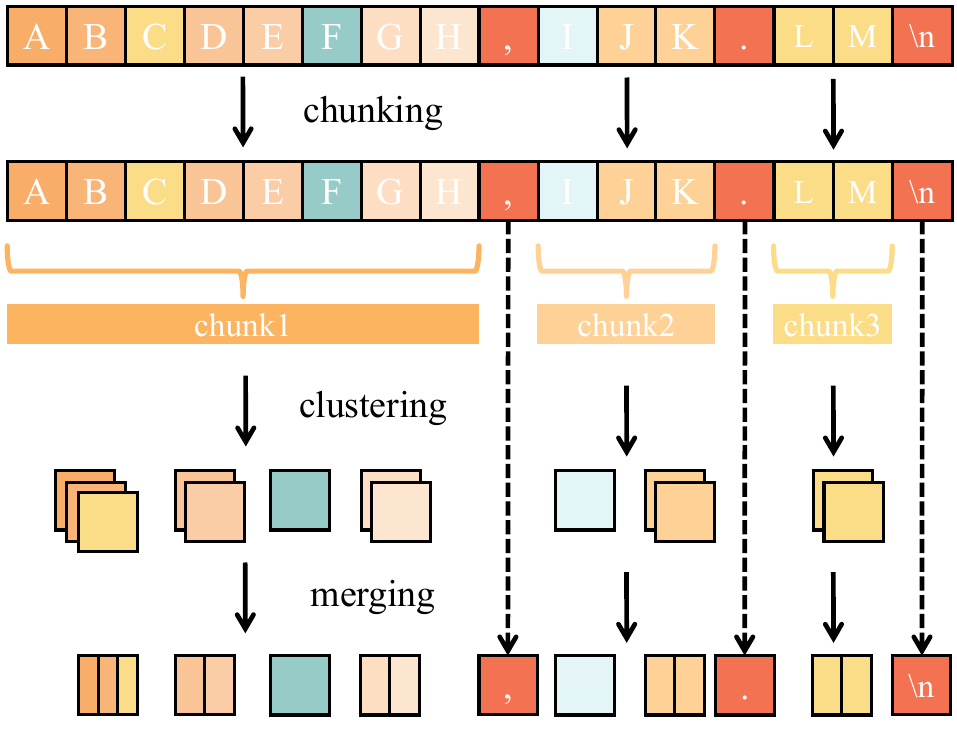}
  \caption{An overview of SemantiCache.}
  \label{fig:method_overview}
\end{figure}

\vspace{-10pt}
\section{Method}
\label{sec:method}

\subsection{Overall Pipeline}
\label{ssec:architecture}

To compress the KV cache while addressing semantic fragmentation, we introduce SemantiCache, a novel three-stage, semantically hierarchical pipeline. As depicted in Figure~\ref{fig:method_overview}, our approach begins with semantic chunking, where the original KV cache is segmented into semantically coherent chunks using natural language delimiters as boundaries. Subsequently, a similarity clustering stage is performed on each chunk, employing the GSC algorithm to group highly similar tokens into distinct sets, which we refer to as semantic clusters. In the final clustered merging stage, the Key-Value (KV) states within each cluster are consolidated into what we define as semantic cores. The compressed KV cache is then constructed by concatenating these semantic cores with KV states of the original delimiters preserved from the initial chunking phase, effectively reducing cache size while maintaining semantic integrity.

\subsection{Semantic Chunking}
\label{ssec:delimiter-chunking}

The semantic chunking stage partitions the KV cache into semantically coherent chunks, aligning these chunks with natural linguistic boundaries—delimiters. This approach contrasts with prior techniques \cite{yuan-etal-2025-native,lu2025moba} that employ fixed-size or arbitrary chunking, which can break semantic boundaries and result in semantic fragmentation.

Specifically, let the KV cache be denoted as 
$(K, V) = [(k_1, v_1), \dots, (k_n, v_n)]$, where $k_i \in \mathbb{R}^d$ and $v_i \in \mathbb{R}^d$ are the $i$-th key and value vectors of the cache, respectively. 
We predefine a set of delimiter tokens, denoted as $\mathcal{D}$. 
Inspired by the observation that delimiters consistently receive high attention scores \cite{sepLLM2025}, we preserve their corresponding KV states unmodified to act as structural anchors. 
The chunking process partitions the sequence $(K, V)$ into an alternating series of non-delimiter chunks and delimiters. 
Formally, this yields a structured sequence:
\begin{equation}
\label{eq:chunking}
    (C_1, d_1, C_2, d_2, \dots, C_{M-1}, d_{M-1}, C_M),
\end{equation}
where $C_m=\{ (k_r, v_r), \dots, (k_{r+l-1}, v_{r+l-1}) \}$ is the $m$-th semantically coherent chunk of length $l$, $d_m \in \mathcal{D}$ is the $m$-th delimiter.

\subsection{Similarity Clustering}
\label{ssec:sim-clustering}

Following the initial stage, each chunk $C_m$ contains a group of semantically coherent, yet potentially redundant KV states. The purpose of similarity clustering is to partition these chunks into finer-grained semantic units by identifying and grouping tokens that are semantically alike. 

To formally measure the semantic affinity between tokens, we leverage their Key states, which encode rich semantic information. For any two KV states $(k_i, v_i)$ and $(k_j, v_j)$ within a chunk, their semantic similarity is defined as:
\begin{equation}
\label{eq:cosine_similarity}
\text{Sim}((k_i,v_i), (k_j,v_j)) = \frac{k_i \cdot k_j}{\|k_i\|_2 \|k_j\|_2},
\end{equation}
where $\cdot$ denotes the dot product and $\| \cdot \|_2$ is the L2 norm.

Based on this metric, we introduce a computationally efficient \textbf{Greedy Seed-Based Clustering (GSC)} algorithm, formally detailed in Algorithm~\ref{alg:gsc}. GSC operates on each chunk from Equation~\ref{eq:chunking} independently. It performs a single, sequential pass: an unassigned token is designated as a "seed" for a new cluster. This seed then greedily absorbs all subsequent unassigned tokens whose key vectors exhibit a semantic similarity (Equation~\ref{eq:cosine_similarity}) exceeding a predefined threshold, $\tau$. This process repeats, with the next available unassigned token forming a new seed, until all KV states in the chunk have been assigned to exactly one cluster. This deterministic, single-pass approach partitions the tokens into disjoint clusters, each exhibiting high internal semantic cohesion.

\begin{algorithm}[h!]
\caption{Greedy Seed-Based Clustering (GSC)}
\label{alg:gsc}
\begin{algorithmic}[1]

\Require
    \begin{varwidth}[t]{\linewidth}
        chunk $C_{m} = \{(k_r, v_r), \dots, (k_{r+l-1}, v_{r+l-1})\}$. \\
        semantic similarity threshold $\tau$.
    \end{varwidth}
\Ensure
    \begin{varwidth}[t]{\linewidth}
        A partition $\mathcal{P}_m = \{S_1, \dots, S_p\}$ of chunk $C_m$ , \\ where each cluster $S_k$ is a set of KV states.
    \end{varwidth}

\State $\textit{assigned} \leftarrow \text{array of size } $r + l$ \text{, initialized to \textbf{False}}$
\State $\mathcal{P}_m \leftarrow \emptyset$
\For{$i = r, \dots, r+l-1$}
    \If{$\textit{assigned}[i]$ is \textbf{False}}
        \State $\text{cluster} \leftarrow \{(k_i, v_i)\}$
        \State $\textit{assigned}[i] \leftarrow \textbf{True}$
        \State $\textit{seed} \leftarrow (k_i,v_i)$
        \For{$j = i + 1, \dots, r+l-1$}
            \If{$\textit{assigned}[j]$ is \textbf{False}}
               \If{$\textproc{Sim}(\textit{seed}, (k_j,v_j)) > \tau$}
                    \State $\text{cluster} \leftarrow \text{cluster} \cup \{(k_j, v_j)\}$
                    \State $\textit{assigned}[j] \leftarrow \textbf{True}$
                \EndIf
            \EndIf
        \EndFor
        \State $\mathcal{P}_m \leftarrow \mathcal{P}_m \cup \{\text{cluster}\}$
    \EndIf
\EndFor
\State \Return $\mathcal{P}_m$
\end{algorithmic}
\end{algorithm}

\subsection{Clustered Merging}
\label{ssec:compression}

Upon obtaining the semantic clusters, the final stage is clustered merging. A naive merge operation would dilute the attention contribution of the consolidated tokens. To counteract this, we propose a two-part solution: creating representative vectors via mean-pooling and applying a \textbf{Proportional Attention} mechanism to reweight their influence.

First, for each cluster $S_t$ within the partition $\mathcal{P}_m$ produced by Algorithm~\ref{alg:gsc}, we compute a single representative semantic core $(\hat{k}_t, \hat{v}_t)$ via mean-pooling, formulated as follows:
\begin{equation}
\label{eq:merging}
\hat{k}_t = \frac{1}{|S_t|} \sum_{(k_i, v_i) \in S_t} k_i \quad \text{and} \quad \hat{v}_t = \frac{1}{|S_t|} \sum_{(k_i, v_i) \in S_t} v_i,
\end{equation}
where $|S_t|$ denotes the size of $S_t$. The final compressed KV cache, denoted $(K_c, V_c)$, is constructed by concatenating these semantic cores with the KV states of the delimiters preserved from the chunking stage.

Subsequently, we apply a modified attention mechanism to rebalance the reduced attention contributions of the merged tokens:
\begin{equation}
\label{eq:proportional_attention}
\text{Attention}(Q, K_c, V_c) = \text{softmax}\left(\frac{Q K_c^T}{\sqrt{d}} + \log \mathbf{s}\right) V_c,
\end{equation}
where $\mathbf{s}$ is a row vector containing the size of each cluster (e.g., $|S_t|$) and $Q \in \mathbb{R}^{n \times d}$ is the query matrix. Mathematically, adding $\log \mathbf{s}$ to the pre-softmax logits is equivalent to multiplying the attention scores by $\mathbf{s}$ after the exponential operation. This ensures that a merged token's contribution is proportional to its original size, effectively preventing the information dilution that would otherwise occur. This final stage significantly reduces the KV cache size while preserving the semantic integrity of the original context.

\section{Experiments}
\label{sec:experiments}

\subsection{Experimental Settings}
\label{ssec:Experimental Settings}

\textbf{Tasks.} We evaluate our method on a diverse set of long-context tasks. We use \textbf{LongBench} \cite{bai-etal-2024-longbench}, 
a multi-task benchmark including 21 datasets designed to test a model's long-context understanding across single-document QA, multi-document QA, summarization, few-shot learning, synthetic tasks, and code completion. Additionally, we use the \textbf{Needle-in-a-Haystack (NIAH)} \cite{kamradt2024needle} test to specifically measure the ability of the model to retrieve a specific sentence within a large document.

\textbf{Models.} We implement SemantiCache on several popular open-source LLMs to demonstrate its generalizability, including \textbf{Llama-3-8B-Instruct} \cite{grattafiori2024llama} and \textbf{Mistral-7B-Instruct-v0.2} \cite{jiang2023mistral7b}.

\textbf{Baselines.} We evaluate SemantiCache against a range of state-of-the-art KV cache compression methods. The selected baselines include eviction-based approaches, such as StreamingLLM \cite{xiao2023efficient}, H2O \cite{zhang2023h2o}, and SnapKV \cite{li2024snapkv}, and merging-based techniques, including CaM \cite{zhang2024cam} and D2O \cite{wan2024d2o}.

\textbf{Implementation Details.} All experiments are performed on NVIDIA A100 80GB GPUs. The delimiter set $\mathcal{D}$ used for Semantic Chunking is defined as [".", ",", "?", "!", ";", ":", " ", "\textbackslash{}t", "\textbackslash{}n"]. We select the $\tau$ value for the GSC algorithm from the range [0.5, 0.9].

\subsection{Accuracy Evaluation}
\label{ssec:Accuracy Evaluation}

\textbf{LongBench Results.} As shown in Table~\ref{tab:longbench_results}, SemantiCache consistently outperforms all other compression baselines on LongBench tasks across various KV cache budgets and models, including both eviction-based and merging-based approaches. This demonstrates that by preserving semantic integrity, SemantiCache effectively mitigates the performance decline typically associated with high compression ratios.

\begin{table}[h!]
\centering
\caption{Average scores on LongBench for different compression methods, evaluated across various cache budgets and models.}
% \caption{LongBench results.}
\label{tab:longbench_results}
\small
\setlength{\tabcolsep}{4pt} % Reduces space between columns to make the table fit
\begin{tabular*}{\columnwidth}{@{\extracolsep{\fill}}l|ccc|ccc}
\toprule
\multirow{2}{*}{\textbf{Method}} & \multicolumn{6}{c}{\textbf{Model}} \\
\cmidrule(lr){2-7}
& \multicolumn{3}{c|}{\textbf{Llama-3-8B}} & \multicolumn{3}{c}{\textbf{Mistral-7B}} \\
\midrule
Full Model& \multicolumn{3}{c|}{34.88} & \multicolumn{3}{c}{42.01} \\
\midrule
& \textbf{20\%} & \textbf{35\%} & \textbf{50\%} & \textbf{20\%} & \textbf{35\%} & \textbf{50\%} \\
\cmidrule(lr){2-7}
StreamingLLM & 27.88 & 28.21 & 29.16 & 28.27 & 30.14 & 31.26 \\
H2O & 28.73 & 30.45 & 30.88 & 35.35 & 37.69 & 37.95 \\
SnapKV & 27.22 & 30.27 & 30.35 & 36.34 & 38.22 & 38.86 \\
CaM & 27.86 & 29.52 & 29.97 & 35.27 & 37.54 & 38.21 \\
D2O & 29.23 & 31.22 & 31.86 & 39.05 & 39.56 & 40.02 \\
\textbf{Ours} & \textbf{30.01} & \textbf{32.17} & \textbf{32.34} & \textbf{39.68} & \textbf{40.09} & \textbf{40.87} \\
\bottomrule
\end{tabular*}
\end{table}
\textbf{Needle-in-a-Haystack Results.} We test our method using Mistral-7B-Instruct-v0.2 \cite{jiang2023mistral7b} on contexts with maximum lengths of 8k and 32k. The KV cache budget was set to 1024 and 4096. The average accuracy is reported in Table~\ref{tab:NIAH}. Our method outperforms all other compression baselines. This demonstrates our method's robust long-context retrieval capabilities, even with a compressed KV cache.

\begin{table}[h!]
\small
\centering
\caption{Needle-in-a-Haystack results.}
\setlength{\tabcolsep}{6pt} % 减小列间距，默认值通常是 6pt
\begin{tabular}{lcccc}
\toprule
\textbf{Methods} & \textbf{L=8k} & \textbf{L=32k} & \textbf{L=8k} & \textbf{L=32k} \\
\midrule
Full Model & 97.56 & 93.85 & 97.56 & 93.85 \\
\midrule
& \multicolumn{2}{c}{\textbf{1024}} & \multicolumn{2}{c}{\textbf{4096}} \\
\midrule
StreamingLLM & 57.42 & 46.81 & 61.63 & 50.88 \\
H2O & 78.96 & 68.21 & 81.62 & 71.88 \\
SnapKV & 82.67 & 75.84 & 85.92 & 80.29 \\
CaM & 81.78 & 77.33 & 86.52 & 77.20 \\
D2O & 90.38 & 86.82 & 93.21 & 90.29 \\
\midrule
\textbf{Ours} & \textbf{91.02} & \textbf{87.23} & \textbf{94.38} & \textbf{91.15} \\
\bottomrule
\end{tabular}
\label{tab:NIAH}
\end{table}

\subsection{Efficiency Evaluation}
\label{ssec:Efficiency Evaluation}

We evaluate the efficiency of our method under a 20\% KV cache budget, focusing on inference latency and memory footprint. We measure latency using two key metrics: \textbf{Time To First Token (TTFT)}, which measures the prefill latency, and \textbf{Time Per Output Token (TPOT)}, which measures the decoding latency. As shown in Table~\ref{tab:efficiency}, SemantiCache achieves a state-of-the-art \textbf{2.61$\times$} speedup in TPOT over the Full KV cache baseline. Furthermore, our approach exhibits a more favorable TTFT compared to other merging-based baselines, underscoring the efficiency of the GSC algorithm. In terms of memory footprint, it outperforms other compression methods, demonstrating its effectiveness in optimizing resource utilization.

\begin{table}[h!]
\small
\centering
\caption{Efficiency evaluation results on Llama-3-8B with a 32k context length.}
\setlength{\tabcolsep}{5pt}
\label{tab:efficiency}
\begin{tabular}{lccc}
\toprule
\textbf{Method} & \textbf{TTFT (s)} & \textbf{TPOT (s)} & \textbf{Memory (GB)} \\
\midrule
Full Model & \textbf{4.12} & 0.081 & 24.27 \\
\midrule
StreamingLLM & \textbf{4.12} & 0.032 & 16.12 \\
H2O & \textbf{4.12} & 0.033 &  16.58\\
SnapKV & 4.13 & 0.032 & 16.37 \\
CaM & 4.28 & 0.039 & 17.03 \\
D2O & 4.29 & 0.038 & 16.91 \\
\textbf{Ours} & 4.25 & \textbf{0.031} & \textbf{15.94} \\
\bottomrule
\end{tabular}
\end{table}

\subsection{Ablation Study}
\label{ssec:ablation_study}

\textbf{Importance of Semantic Chunking.}
To validate the efficacy of our Semantic Chunking stage, we perform an ablation study comparing the full SemantiCache pipeline against two baselines: (1) a variant without any chunking, where Similarity Clustering is applied directly to the entire token sequence, and (2) a variant employing a standard fixed-size chunking strategy (64-token blocks). Performance is evaluated  on the LongBench average score under a 20\% KV cache budget using Llama-3-8B. As shown in Table~\ref{tab:ablation_chunking}, our Semantic Chunking achieves superior performance, which we attribute to its ability to preserve the integrity of complete linguistic units.

\begin{table}[h!]
\centering
\caption{Comparison of different chunking strategies.}
\label{tab:ablation_chunking}
\begin{tabular}{lc}
\toprule
\textbf{Method Variant} & \textbf{Average Score} \\
\midrule
\textbf{Ours (Semantic Chunking)} & \textbf{30.01} \\
Fixed-size Chunking (64 tokens) & 28.22 \\
w/o Chunking (Global Clustering) & 26.94 \\
\bottomrule
\end{tabular}
\end{table}

\textbf{Effect of Semantic Similarity Threshold $\tau$.}
The semantic similarity threshold, $\tau$, is the key hyperparameter in our GSC algorithm, controlling the balance between compression ratio and model performance. We conduct experiments on different $\tau$ values from the range [0.5, 0.9] on LongBench using Llama-3-8B-Instruct. As shown in Table~\ref{tab:ablation_threshold}, a lower $\tau$ leads to more aggressive compression at the cost of performance. Conversely, a higher $\tau$ preserves accuracy by creating more, smaller clusters, thus reducing the compression rate.

\begin{table}[h!]
\small
\centering
\caption{Comparison of different values for $\tau$.}
\setlength{\tabcolsep}{3pt}
\label{tab:ablation_threshold}
\begin{tabular}{ccc}
\toprule
\textbf{Threshold ($\tau$)} & \textbf{Average Score} & \textbf{Compression Ratio (\%)} \\
\midrule
0.50 & 27.21 & 86.3 \\
0.60 & 28.46 & 82.5 \\
0.70 & 30.12 & 79.8 \\
0.80 & 32.87 & 51.2 \\
0.90 & 34.05 & 16.7 \\
\bottomrule
\end{tabular}
\end{table}

\section{Conclusion}
\label{sec:conclusion}

In this study, we introduce SemantiCache, a novel framework designed to address the critical issue of semantic fragmentation prevalent in existing KV cache compression techniques. Our method aligns the compression process with the natural hierarchical structure of language, utilizing a coarse-to-fine-grained, three-stage pipeline that preserves the semantic integrity often destroyed by conventional approaches. Extensive experiments demonstrate that SemantiCache not only yields significant gains in inference speed and memory efficiency but also substantially outperforms state-of-the-art eviction and merging methods on various long-context tasks. 
%\vfill
%\pagebreak

\bibliographystyle{IEEEbib}
\bibliography{refs}

\end{document}